\documentclass[10pt, conference, compsocconf]{IEEEtran}
\pdfoutput=1

\IEEEoverridecommandlockouts
\usepackage{cite}
\usepackage{amsmath,amssymb,amsfonts}
\usepackage{algorithmic}
\usepackage{tikz}
\usetikzlibrary{shapes.geometric, arrows}
\usepackage{graphicx}
\usepackage{textcomp}
\usepackage{xcolor}
\def\BibTeX{{\rm B\kern-.05em{\sc i\kern-.025em b}\kern-.08em
    T\kern-.1667em\lower.7ex\hbox{E}\kern-.125emX}}

\usepackage{color,soul}
\usepackage{todonotes}
\usepackage{multirow}
\usepackage{tabularx}
\usepackage{enumerate}
\usepackage[caption=false,font=small]{subfig}
\usepackage{framed}
\usepackage{color}
\usepackage{listings}
\definecolor{color1}{rgb}{0,0,1}
\definecolor{color2}{rgb}{1,0,0}
\definecolor{color3}{rgb}{0,1,0}
\definecolor{color4}{rgb}{0.5,0.5,0}
\definecolor{color5}{rgb}{0,0.5,0.5}
\definecolor{color6}{rgb}{0.5,0,0.5}
\definecolor{color7}{rgb}{0.5,0.5,0.5}
\usepackage{wrapfig}
\usepackage[hyphens]{url}
\usepackage[hidelinks]{hyperref}
\hypersetup{breaklinks=true}
\usepackage{cite}
\usepackage{xspace}      
\definecolor{emphcolor}{rgb}{0.50,0.10,0.10}
\usepackage[lined, linesnumbered, commentsnumbered]{algorithm2e}
\usepackage{etoolbox}
\makeatletter
\patchcmd{\algocf@latexcaption}{#3}{#3\endgraf}{}{}
\SetAlgoCaptionSeparator{.\ \ \ \ }
\SetAlCapNameFnt{\footnotesize}
\SetAlCapFnt{\footnotesize}

\usepackage{nicefrac}
\usepackage{tablefootnote}

\newcommand{\OpenCL}{OpenCL\textsuperscript{\textregistered} }
\newcommand{\Python}{Python\texttrademark{}~}
\begin{document}

\title{MIOpen: An Open Source Library For \\Deep Learning Primitives \\
}

 \author{\IEEEauthorblockN{Jehandad Khan, Paul	Fultz, Artem Tamazov, Daniel	Lowell, Chao	Liu,\\ Michael	Melesse, 
 Murali	Nandhimandalam, Kamil	Nasyrov, Ilya	Perminov, Tejash	Shah,\\Vasilii Filippov, Jing	Zhang, Jing	Zhou, Bragadeesh Natarajan, Mayank Daga} 
 \IEEEauthorblockA{\textit{AMD Inc.} \\
 Mayank.Daga@amd.com}}
\maketitle

\begin{abstract}
Deep Learning has established itself to be a common occurrence in the business lexicon. The unprecedented success of deep learning in recent years can be attributed to: abundance of data, availability of gargantuan compute capabilities offered by GPUs, and adoption of open-source philosophy by the researchers and industry. Deep neural networks can be decomposed into a series of different operators. MIOpen, AMD's open-source deep learning primitives library for GPUs, provides highly optimized implementations of such operators, shielding researchers from internal implementation details and hence, accelerating the time to discovery. This paper introduces MIOpen and provides details about the internal workings of the library and supported features. 

MIOpen innovates on several fronts, such as implementing fusion to optimize for memory bandwidth and GPU launch overheads, providing an auto-tuning infrastructure to overcome the large design space of problem configurations, and implementing different algorithms to optimize convolutions for different filter and input sizes. MIOpen is one of the first libraries to publicly support the bfloat16 data-type for convolutions, allowing efficient training at lower precision without the loss of accuracy.
\end{abstract}


\section{Introduction}
Deep Learning has burgeoned into one of the most important technological breakthroughs of the 21st century. The use of deep learning has garnered immense success in applications like image and speech recognition, recommendation systems, and language translation. This in turn advances fields like autonomous driving and disease diagnosis \cite{zhang2019reducing}. GPUs have played a critical role in the advancement of deep learning. The massively parallel computational power of GPUs has been influential in reducing the training time of complex deep learning models hence, accelerating the time to discovery \cite{you2018imagenet}. The availability of open-source frameworks like TensorFlow and PyTorch is another cornerstone for the fast-paced innovation in deep learning~\cite{abadi2016tensorflow, paszke2017pytorch}.

The deep learning frameworks decompose the models as either a computational graph or a sequence of operations~\cite{jia2014caffe, mlir, Cyphers2018}. These high-level operations are then compiled down to a series of hardware specific high-performance primitives. These primitives in deep learning are akin to BLAS (Basic Linear Algebra Subprograms)~\cite{lawson1977basic} in linear algebra and high performance computing. Availability of a library which provides highly optimized implementations of such primitives enables the deep learning researchers to focus on their science and leaves the burden of developing such primitives on the hardware vendors. The library then provides a simple and callable application programming interface (API) to enable seamless integration with client libraries and be flexible so that new features may be added easily.

MIOpen is AMD's deep learning primitives library which provides highly optimized, and hand-tuned implementations of different operators such as \emph{convolution}, \emph{batch normalization}, \emph{pooling}, \emph{softmax}, \emph{activation} and layers for \emph{Recurrent Neural Networks (RNNs)}, used in both training and inference~\cite{hochreiter1997lstm}. Moreover, MIOpen is fully open-source including all its GPU kernels; complementing AMD's open-source ROCm stack~\cite{rocm}. MIOpen is the \textit{first} to extend the open-source advantage into GPU vendor libraries thereby, continuing to embark on the same ethos as the deep learning community.

As deep learning has gained critical acclaim over the years, substantial research has been conducted to accelerate it. One optimization technique called \textit{fusion} has been recognized to be more potent than others\cite{leary2017xla}. Fusion allows to fuse or collapse several neural network layers thereby, optimizing on 1) memory bandwidth requirements by requiring less data to be moved between host and GPU memories, and 2) GPU kernel launch overheads by launching fewer GPU kernels compared to the vanilla, non-fused neural network. Aside from discrete primitives, MIOpen also offers a fusion API which allows the frameworks to fuse some of the operations mentioned above. MIOpen fusion can be used to accelerate both convolution and recurrent neural networks.

Another area that has flourished with the popularity of deep learning is open-source graph compilers \cite{leary2017xla}, \cite{rotem2018glow}, \cite{lattner2019mlir}, \cite{chen2018tvm}. Graph compilers further the relevance of deep learning to wide-spread applications by generating the implementations of aforementioned operators instead of relying on hardware specific libraries. However, generating high-performance implementations of two operators, convolution and GEMM, is extremely cumbersome without inherent knowledge of the underlying hardware. Therefore, the graph compilers rely on libraries like MIOpen for these operators. MIOpen's open-source nature enables a plethora of optimization opportunities which were not possible before. For example, fusing an operator generated by the compiler with MIOpen's convolutions. MIOpen facilitates these optimization by breaking down complex operators like convolutions into several simple and small operators and providing high-performance implementations of these simple operators to the graph compiler. This MIOpen feature is called \textit{composable~kernels}.

The primary aim of MIOpen is to provide access to high-performance kernels, support several data-types, and also support as many hardware targets as required. To that end, MIOpen supports four different data-types: \texttt{float32, float16, bfloat16, and int8}, and two programming models: HIP and \OpenCL~\cite{munshi2009opencl}. The kernels in MIOpen are backed by both high-level language as well as hand-tuned assembly implementations. MIOpen also provides an auto-tuning infrastructure to achieve maximum performance on the user's hardware and software environment.



This document provides an under-the-hood look at the MIOpen library providing detailed information about the functionality of the library as well as introduce MIOpen's capabilities to users and developers. The rest of the paper is organized as follows: Section \ref{sec_prior} describes some prior work, Section \ref{sec_design} describes the overall design philosophy of the library and provides details about kernel compilation, abstractions used to localize those details in the library, tuning infrastructure for improving kernel performance and MIOpen's support for \OpenCL and HIP. This is followed by Section \ref{sec_prims} which provides details about the supported operations; primarily the convolution operation. Section \ref{sec_fusion} describes MIOpen's Fusion API for merging different operations for increased performance, this is followed by some usage statistics and performance comparisons in Section \ref{sec_stat}. Section \ref{sec_conc} presents conclusion and future work.

\section{Related Work}
\label{sec_prior}
Developing hardware-optimized libraries for most critical and time-sensitive operations is a well-known practice.
For linear algebra such libraries are known as BLAS (Basic Linear Algebra Subsystem) and have different implementations for different systems~\cite{lawson1977basic, whaley1998automatically, belter2009automating, frigo1998fftw}. In similar spirit different deep learning libraries have been written, to make it easier for client applications to implement different deep learning primitives. Alex Krischevsky's \texttt{cuda-convnet} is one of the initial libraries to implement convolutions and inspired many others~\cite{Vasilache2018}, \cite{mkldnn}. 
Chetlur et al.\ developed cuDNN, a deep neural network library for nVIDIA GPUs~\cite{Chetlur2014}. MIOpen falls in this category since it provides a C programming language based API for deep learning primitives. While these libraries aim to accelerate deep learning primitives on GPUs, research also been conducted to improve the performance of inference only loads on different CPUs such as MKL-DNN~\cite{mkldnn}.

Most of the above mentioned libraries focus on lower level optimization opportunities. An orthogonal approach is to abstract this detail behind a domain specific language (DSL). This technique has already been successfully applied to other domains such as computer vision and linear algebra~\cite{ragan2013halide, kjolstad2017tensor, kjolstad2016simit, lujan2000oolala}.  Vasilache et al.\ developed \textit{Tensor Comprehensions}, which takes a similar approach and designs a language which can infer tensor dimensions and summation indices automatically~\cite{Vasilache2018}. However, such an approach makes it complicated to support a wide array of platforms and hardware targets as is required of MIOpen. 

\subsection{MIOpen and higher level frameworks}
The above libraries are augmented by a community of frameworks which enable researchers and practitioners to express their computation pipeline using a host language (typically \Python or some other higher level language)\cite{jia2014caffe}\cite{abadi2016tensorflow}\cite{Chen2015}\cite{collobert2012implementing}\cite{al2016theano}. These frameworks in turn call out libraries such as MIOpen for efficient implementation of the primitives required to implement the computation in those graphs. Frameworks strive to support a wide array of hardware and applications, for instance both TensorFlow and PyTorch already support MIOpen as a backend aimed at AMD GPUs. Thus a user can seamlessly change the hardware target without changing their application code. 



\section{Overall Design}
\label{sec_design}
This section describes the MIOpen's design philosophy using the convolution operation as an example.

\subsection{Kernels and Solvers}
Mapping a problem description to a particular kernel requires MIOpen to determine the file which contains the required GPU kernel, the name of the kernel in the file and the compiler arguments required to compile it. Typically, there is more than one kernel which can perform similar operations. However, each kernel has a unique set of constraints and may result in different performance due to differing code optimizations and input dimensions of the problem. For example, one kernel might be the best choice for large image sizes while another may perform better for smaller image sizes, each using different coding patterns for optimum performance.

All this information is grouped in MIOpen classes collectively called \emph{solvers}. These classes together \emph{solve} for the best convolution kernel given a problem description. This construct creates a layer of abstraction between the rest of the MIOpen library and kernel specific details, thus all the details of a kernel are completely localized. A solver is trivially constructible by design and therefore has no state, this ensures that kernel compilation launches do not have side effects.

If a developer wishes to add a new kernel to the MIOpen, all that is required is to add the source code for the kernel and implement the associated solver, thereafter the kernel may be selected automatically.

\subsection{Auto tuning infrastructure}
In general, any high-performance code leverages auto-tuning for choosing the parameters that may change with the underlying architecture as well as the problem description thereby, impacting performance. MIOpen is not an exception to this rule. This requires that all tunable kernels be tuned for known configuration to achieve maximum performance. Once known, these tuning parameters can be shipped with MIOpen or, the user may employ the same infrastructure to tune MIOpen kernels for custom configurations. 

A solver encapsulates the constraints for the tuning parameters as well as the interface machinery to launch tuning instances. The tuning parameters create a grid of possible values of the kernel tuning parameters and the tuning infrastructure compiles and launches a unique kernel for each of these combinations using a pruned search space approach. Once a kernel is tuned and the optimum tuning parameters are known, they are serialized to a designated directory on the user's system for future retrieval.

MIOpen ships with optimized tuning parameters for many configurations used in popular convolutional neural networks. Moreover,  the user may run tuning sessions to further optimize kernel codes on their hardware or to add configurations for specific use cases. Further details about the tuning process can be found at \cite{miopen_doc}.

\subsection{Kernel compilation and caching}
Launching a kernel requires setting up the compilation parameters and invoking a device-code compiler to generate the binary object. MIOpen device-code consists of kernels written in \OpenCL, HIP\cite{amdhip} and GCN assembly\cite{gcnisa}, which may be compiled using \emph{clang}\cite{clang:URL}.

Since compiling a kernel is a costly and time-consuming procedure, MIOpen employs two levels of caching to improve the runtime performance of the library. This design choice is tightly coupled with how device-code compilers  compile and load compute-kernels from the binaries.

Once an kernel file is compiled, it is cached to disk to avoid future compilations of the same source-file with the same parameters. The specific kernel that would be invoked is loaded into memory from the disk and stored in an in-memory cache for subsequent invocation by the same program. This results in substantial runtime improvements since neural network models typically invoke the same kernels many times during an application's lifetime. 

Due to the caching effects described above, it is recommended that the user's application performs a \emph{warmup} iteration so that MIOpen's different caches can be populated. Such runs will ensure that subsequent network invocations are accurately timed without the effects of disk I/O or compilation delays. This limitation is not unique to MIOpen and is also applicable to other high-performance libraries.

\subsection{ HIP and \OpenCL backends}
\begin{figure}[hbt]
	\centering
	{\includegraphics[scale=0.6]{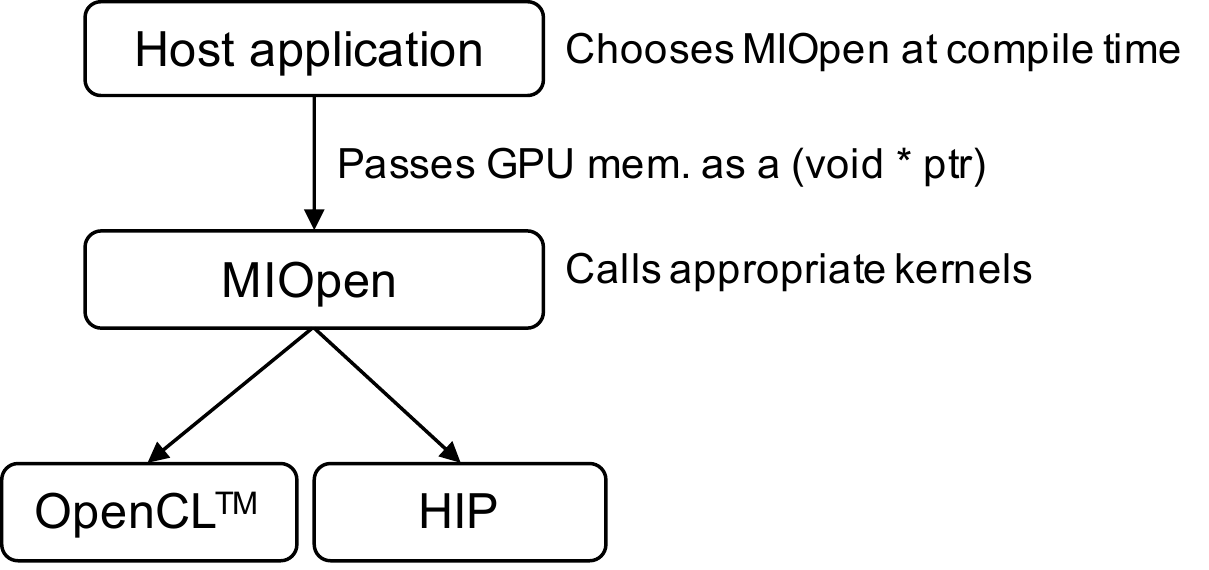}}
	\caption{MIOpen supports \OpenCL and HIP Programming Models}
	\label{fig_backend}
\end{figure}

MIOpen supports applications that use the \OpenCL and HIP programming models~\cite{amdhip}. 
As shown in Figure~\ref{fig_backend}, all the APIs remain consistent from the client application's perspective, the only difference is in the creation of \texttt{miopenHandle} structure, which is created either with a HIP stream or an \OpenCL device context. Internally the HIP backend compiles the kernel using an appropriate complier depending on the kernel source type. Subsequently, the compiled binary object is loaded and passed off to the runtime for execution.

%

\section {Machine Learning Primitives}
\label{sec_prims}
\subsection{Convolution}
Most modern neural networks employ convolution as a central operation\cite{lecun1995convolutional}. Its usefulness and popularity make it a critical piece of the machine learning puzzle, particularly in image processing.

Convolution implementations have a large design space due to its numerical complexity and its diverse inputs make it difficult to generalize across multiple architectures. Over the past few years, different algorithms have been proposed to efficiently compute the convolution of an image with a group of filters. Among these, the Winograd algorithm is notable\cite{Lavin2015}. The Winograd algorithm achieves the highest efficiency for some key filter sizes. MIOpen's winograd implementation also provides the benefit of not requiring additional ``workspace'' for intermediate computations. The most general and arguably most expensive in terms of additional storage requirement is to convert the image matrix to a circulant matrix (popularly known as the \textit{im2col} operation \cite{jia2014caffe}), thereafter multiplying the image matrix and the circulant matrix.

Large filter sizes use Fast Fourier Transform (FFT) to convert an image and a filter (after suitable padding) to the frequency domain and then apply a point-wise multiplication followed by the inverse transform to recover the result in time-domain. While this incurs a transformation overhead for the image each time, there are certain cases where this approach is faster than other methods since the filter needs to be transformed only once.

In addition to the above algorithms, MIOpen also implements specialized kernels which directly perform the convolution operation using optimized GCN assembly \cite{gcnisa} or \OpenCL code\cite{munshi2009opencl}. These kernels are collectively known as the \emph{direct} algorithm. 

The best performing algorithm is rarely readily apparent on a given architecture for a set of input and filter dimensions. To assess the relative performance of these kernels and return the best performing kernel, MIOpen employs the \emph{find step} before the actual convolution operation. For this step, the user constructs the necessary data structures for the input/output image tensors as well as the convolution descriptor specifying the properties of convolution such as striding, dilation, and padding. The user then calls the MIOpen convolution \texttt{Find} API which allows MIOpen to benchmark all the applicable kernels for the given problem configuration, this information is returned in an array of type \texttt{miopenConvAlgoPerf\_t}. This enables the library to adjust for any variations in the user hardware and also allows the user to balance the trade-off between execution time and additional memory that may be required for some algorithms. 

The \texttt{miopenConvAlgoPerf\_t} structure mentioned above contains the name of the applicable algorithm, the estimated execution time and the amount of additional memory required by the algorithm. The user may examine this data structure to choose the best algorithm implementation for the problem at hand. This procedure is intended to be performed once and the same find result may be used in subsequent invocations, amortizing its cost. 
\subsection*{Types of convolution}

\subsubsection*{Transpose Convolution}
Transposed Convolution (also known as deconvolution or fractionally-strided convolution) is an operation typically used to increase the size of the tensor resulting from convolution~\cite{dumoulin2016guide}. The standard convolution operation reduces the size of the image, which is desirable in classification tasks. However, tasks such as image segmentation\cite{long2015fully} require the output tensor to have the same size as the input. MIOpen supports transpose convolution required by such networks and may be enabled by setting the \texttt{miopenConvolutionMode\_t} in \texttt{miopenConvolutionDescriptor\_t} to \texttt{miopenTranspose}.

\subsubsection*{Depthwise convolution}
In depthwise convolution, the input is separated along the depth (channels) and then is convolved with a filter that is also separated along the same axis. The results are stacked into a tensor. To understand why this is useful we have to consider the context in which depthwise convolution occurs -- depthwise separable convolution~\cite{sifre2014rigid}. Depthwise separable convolution involves performing a depthwise convolution followed by a 1x1 convolution on the output tensor called a pointwise convolution\cite{howard2017mobilenets}. Separating out the process of finding spatial correlation and cross channel correlations, results in fewer parameters as compared to regular convolution\cite{chollet2017xception}. Smaller and more efficient neural networks with depthwise separable convolutions have applications in training on embedded systems such mobile phones. 

\subsubsection*{Grouped convolutions}
Group convolutions were introduced in Alexnet\cite{krizhevsky2012imagenet}, to reduce the memory required for convolution operation. Grouped convolutions are able to achieve accuracy similar to non-grouped convolutions while having fewer parameters. Moreover, grouped convolutions have a higher level of parallelism \cite{ioannou2017deep}. Conceptually they are a generalization of depthwise convolution, but instead split the input into individual channels and convolve with a filter that is split up the same way. The results are then stacked together. In grouped convolution the input is split up into groups along the channel axis and is then convolved with a filter that has been grouped along the same axis with the output formed by stacking the resulting tensors from each group; further details may be found in~\cite{krizhevsky2012imagenet}.

 To perform a groupwise convolution use the function \texttt{miopenSetConvolutionGroupCount} on a \texttt{miopenConvolutionDescriptor\_t} to set the group count. To perform a depthwise convolution use the same function to set group count to the number of channels~\cite{howard2017mobilenets}.

\subsubsection*{Composable Kernels}
Different variations of the convolution operation discussed above as well as the variety of algorithms that may be used to implement them make it difficult to develop efficient kernels. One solution to tackle this complexity is to break down these operations into reusable modules that can be universally used by different implementations of different algorithms, and express a kernel as a composition of these modules. 

Development work would fall into one of the following categories: 1) describe an algorithm with a hardware-agnostic expression, 2) decide how to map the hardware-agnostic expressions into hardware-dependent modules, 3) implement and optimize the hardware-dependent modules for specific hardware. A potential benefit of breaking down these primitives into smaller modules, is that it then open new doors to optimization that may fuse these modules together.

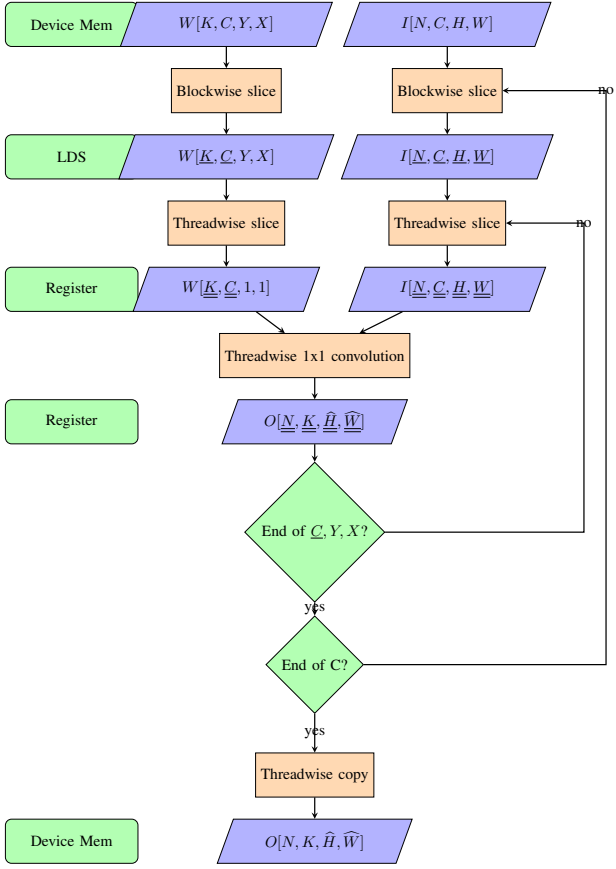
\begin{figure}
    \centering
\resizebox{0.45\textwidth}{!}{%
\begin{tikzpicture}[node distance=1.5cm]
\tikzstyle{Tensor} = [trapezium, trapezium left angle=70, trapezium right angle=110, minimum width=1cm, minimum height=1cm, text centered, draw=black, fill=blue!30]
\tikzstyle{Memory} = [rectangle, rounded corners, minimum width=3cm, minimum height=1cm,text centered, draw=black, fill=green!30]
\tikzstyle{Operator} = [rectangle, minimum width=2cm, minimum height=1cm, text centered, draw=black, fill=orange!30]
\tikzstyle{Decision} = [diamond, minimum width=2cm, minimum height=1cm, text centered, draw=black, fill=green!30]
\tikzstyle{Arrow} = [thick,->,>=stealth]
\tikzstyle{wft} = [diamond, minimum width=3cm, minimum height=1cm, text centered, draw=black, fill=green!30]
\node (mem1) [Memory] {Device Mem};
\node (wei1) [Tensor, right of=mem1, xshift=2cm] {$W[K,C,Y,X]$};
\node (in1) [Tensor, right of=mem1, xshift=7cm] {$I[N,C,H,W]$};
\node (op_wei1) [Operator, below of=wei1] {Blockwise slice};
\node (op_in1) [Operator, below of=in1] {Blockwise slice};
\draw [Arrow] (wei1) -- (op_wei1);
\draw [Arrow] (in1) -- (op_in1);
\node (mem2) [Memory, below of=mem1, yshift=-1.5cm] {LDS};
\node (wei2) [Tensor, right of=mem2, xshift=2cm] {$W[\underline{K},\underline{C},Y,X]$};
\node (in2) [Tensor, right of=mem2, xshift=7cm] {$I[\underline{N},\underline{C},\underline{H},\underline{W}]$};
\draw [Arrow] (op_wei1) -- (wei2);
\draw [Arrow] (op_in1) -- (in2);
\node (op_wei2) [Operator, below of=wei2] {Threadwise slice};
\node (op_in2) [Operator, below of=in2] {Threadwise slice};
\draw [Arrow] (wei2) -- (op_wei2);
\draw [Arrow] (in2) -- (op_in2);
\node (mem3) [Memory, below of=mem2, yshift=-1.5cm] {Register};
\node (wei3) [Tensor, right of=mem3, xshift=2cm] {$W[\underline{\underline{K}},\underline{\underline{C}},1,1]$};
\node (in3) [Tensor, right of=mem3, xshift=7cm] {$I[\underline{\underline{N}},\underline{\underline{C}},\underline{\underline{H}},\underline{\underline{W}}]$};
\draw [Arrow] (op_wei2) -- (wei3);
\draw [Arrow] (op_in2) -- (in3);
\node (op_conv) [Operator, below of=wei3, xshift=2cm] {Threadwise 1x1 convolution};
\draw [Arrow] (wei3) -- (op_conv);
\draw [Arrow] (in3) -- (op_conv);
\node (mem4) [Memory, below of=mem3, yshift=-1.5cm] {Register};
\node (out4) [Tensor, right of=mem4, xshift=4cm] {$O[\underline{\underline{N}},\underline{\underline{K}},\underline{\underline{\widehat{H}}},\underline{\underline{\widehat{W}}}]$};
\draw [Arrow] (op_conv) -- (out4);
\node (dec_endof_yx) [Decision, below of=out4, yshift=-1cm] {End of $\underline{C},Y,X$?};
\node (dec_endof_c) [Decision, below of=dec_endof_yx, yshift=-1.5cm] {End of C?};
\draw [Arrow] (dec_endof_yx) -- node {yes} (dec_endof_c);
\draw [Arrow] (dec_endof_yx) -| ([xshift=4.5cm]  dec_endof_yx.east) |- node {no} (op_in2);
\draw [Arrow] (dec_endof_c) -| ([xshift=5cm]  dec_endof_yx.east) |- node {no} (op_in1);
\draw [Arrow] (out4) -- (dec_endof_yx);
\node (op_out4) [Operator, below of=dec_endof_c, yshift=-1cm] {Threadwise copy};
\draw [Arrow] (dec_endof_c) -- node {yes} (op_out4);
\node (mem5) [Memory, below of=mem4, yshift=-8cm] {Device Mem};
\node (out5) [Tensor, right of=mem5, xshift=4cm] {$O[N,K,\widehat{H},\widehat{W}]$};
\draw [Arrow] (op_out4) -- (out5);
\end{tikzpicture}%
}
    \caption{Work division for convolution using composable kernels }
    \label{fig:implicit_gemm}
\end{figure}

This new kernel programming model is referred to as \emph{composable kernels} in MIOpen. MIOpen v2.0 includes an implementation of the \texttt{implicit GEMM} convolution algorithm, using the composable kernel programming approach. Figure~\ref{fig:implicit_gemm} gives an overview of the overall structure of direct convolution implementation using this methodology. Further details about this novel programming paradigm will be published in the future. 

\subsection{Batch Normalization}
Batch normalization is a very successful technique for accelerating deep neural network training. While initially the improved training dynamics were thought to be a result of reduced internal covariate shift recent research has shown that the true impact of batch normalization layers is a smoother loss function surface making it easier for optimization algorithms to converge to an optimum solution~\cite{ioffe2015batch, shimodaira2000improving}.

There are two versions of batch normalization supported in MIOpen: \texttt{Per-activation} and \texttt{Spatial} batch normalization. Per-activation batch normalization is typically positioned after a fully connected layer in a network\cite{ioffe2015batch}. For a batch of input samples $x$ represented by channel $i$, image height and width as $j$ and $k$ respectively, the per activation batch normalization procedure may be described by:
$$
y_{ijk}=\gamma_{ijk} \Bigg(\frac{x_{ijk} - \mu_{ijk}}{\sqrt{\sigma_{ijk}^2 + \epsilon}}\Bigg) + \beta_{ijk}
$$
Where, $y_{ijk}$ is the output, $\mu_{ijk}$ and $\sigma_{ijk}^2$ are the mean and variance respectively, $\epsilon$ is a small value to avoid division by zero and $\beta_{ijk}$ and $\gamma_{ijk}$ are learned parameters. 

Batch normalization for convolution layers is termed \texttt{spatial} in that it learns separate parameters $\gamma_i$ and $\beta_i$ for each channel, such that the same transform is applied to all the activations in a single feature map\cite{ioffe2015batch}. Likewise, the parameters $\gamma_i$ and $\beta_i$ are learned per channel.

Mathematically,
$$
y_{ijk}=\gamma_i \Bigg(\frac{x_{ijk} - \mu_i}{\sqrt{\sigma_i^2 + \epsilon}}\Bigg) + \beta_i
$$

MIOpen supports the batch normalization operation for both training and inference. They all accept the mode parameter from the \texttt{miopenBatchNormMode\_t} enum, Which has two modes \texttt{miopenBNPerActivation}, which does element-wise normalization for a fully connected layer and \texttt{miopenBNSpatial} which does normalization for convolutions layers. There are specific kernels for training, inference and backward pass for both per activation and spatial batch norm. For more information see \cite{miopen_doc} and \cite{miopen_github}.

\subsection{Recurrent Neural Networks}
The concept of recurrent neural networks (RNN) dates back to the 1980s for storage of the neuron states in self organizing neural networks~\cite{hopfied1982neural}\cite{rumelhart1986learning}. In the Naive RNN structure (also known as vanilla RNN), each neuron is fed with information from input and a previously stored state to predict the next state of the neuron. This attribute of RNNs along with their flexibility in layer construction allows them to substitute for Hidden Markov Model (HMM) to solve state transition problems such as speech recognition and handwriting recognition~\cite{graves2008handwriting}\cite{graves2013speech}. However, vanilla RNN are notoriously difficult to train due to gradient vanishing and exploding issues~\cite{hochreiter1997lstm}. As a remedy, long short-term memory (LSTM) structure was introduce which later proved to be effective in sequence to sequence learning~\cite{graves2008handwriting}\cite{graves2013speech}\cite{sak2014acoustic}. 
Modifications to LSTM such as peephole LSTM, Gated Recurrent Units (GRU) and Simple Recurrent Units (SRU) were later introduced and have been widely used for a variety of applications~\cite{gers2000peephole}\cite{cho2014gru}\cite{lei2018sru}.

MIOpen supports three RNN types prevalent in the industry and research: vanilla RNN, LSTM and, GRU and two kinds of activation function for the hidden state of vanilla RNN neuron: Rectified Linear Unit (ReLU) and hyperbolic tangent (Tanh). Furthermore, information through the RNN may flow in the forward direction (unidirectional RNNs) or both in the forward and backward directions (bi-directional RNNs). MIOpen supports all three RNN types in the unidirectional \texttt{miopenRNNunidirection} as well as the bidirectional model \texttt{miopenRNNbidirection}. Some RNN layers take input sequences directly from the output of a previous layer while others require a transform to align the intermediate vector dimension or simply to achieve better results. MIOpen satisfies this requirement by supporting two input types: 1) \texttt{miopenRNNlinear}, which performs a linear transform before feeding the input to the neuron, and 2) \texttt{miopenRNNskip}, which allows a direct input into the neuron. Similarly, bias to the neural network may be added or removed by choosing the mode \texttt{miopenRNNWithBias} or \texttt{miopenRNNNoBias}. 

The dependence of current state on the previous state as well as different RNN configurations make it difficult to achieve high computational efficiency on a GPU platform. Prevalent frameworks such as TensorFlow encapsulate the state updating functions of the RNN neuron in a cell format to achieve better compatibility in different modes, though the impact of the data layouts and computation procedures on performance is neglected~\cite{tfrnncell}. MIOpen handles the RNN computation by taking advantage of two powerful ROCm platform GEMM libraries (1) rocBLAS for the HIP backend, and (2) MIOpenGEMM for the \OpenCL backend, which are augmented by specialized MIOpen kernels for other primitive functions. 

\subsubsection{Fusion and LSTM}
In the following paragraphs, details about  the fusion and optimization of LSTM's forward and backward paths are presented. Similar ideas extend to the design of vanilla RNN as well as GRU in MIOpen. The interested reader is referred to the MIOpen code repository for further details\cite{miopen_github}.

\begin{figure}
	\includegraphics[scale=0.3]{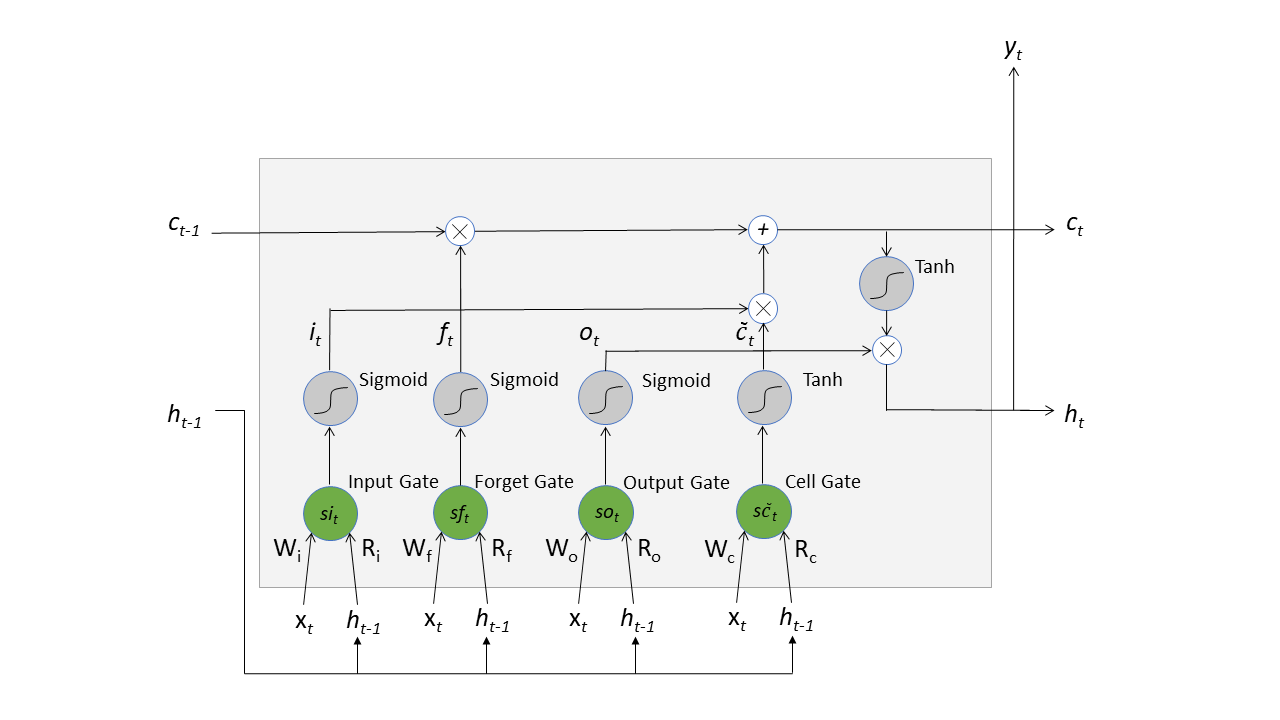}
	\caption{structure of LSTM neuron and forward flow}
	\label{fig_lstm_fwd}
\end{figure}

LSTM has four gates to support its storage and update in long and short-term memories. The structure of LSTM neuron is shown in Figure \ref{fig_lstm_fwd} and its recurrent logic are shown in equation \ref{eq_fwdsi}-\ref{eq_fwdhid}.
		\begin{equation}
si_t=W_ix_t+R_ih_{t-1}
		\label{eq_fwdsi}
		\end{equation}
		\begin{equation}
sf_t=W_fx_t+R_fh_{t-1}
		\label{eq_fwdsf}
		\end{equation}	
		\begin{equation}
so_t=W_ox_t+R_oh_{t-1}
		\label{eq_fwdso}
		\end{equation}	
		\begin{equation}
s\tilde{c}_t=W_{c}x_t+R_{c}h_{t-1}
		\label{eq_fwdsc}
		\end{equation}
In the equations, the subscript $t$ denotes the time index in the sequence, and $i_t$, $f_t$, $o_t$ and $c_t$ denote the updates at the \emph{input}, \emph{forget}, \emph{output} and \emph{cell} gates respectively at time $t$. The input is deonted as $x$, and $h$ indicates the hidden state of the LSTM cell. The matrices $W_i$, $W_f$, $W_o$, $W_c$ are the weight matrices of $x$ for the input, forget, output and cell gates respectively. Similarly, the $R$ matrices with appropriate subscripts are the gain matrices for the hidden state $h$. The above equations represent the linear transformation of the various LSTM states, these interim states go through different activation functions as follows:
		\begin{equation}
i_t=sigmoid(si_t)
		\label{eq_fwdi}
		\end{equation}	
		\begin{equation}
f_t=sigmoid(sf_t)
		\label{eq_fwdf}
		\end{equation}	
		\begin{equation}
o_t=sigmoid(so_t)
		\label{eq_fwdo}
		\end{equation}	
		\begin{equation}
\tilde{c}_t=tanh(s\tilde{c}_t)~~~~
		\label{eq_fwdc}
		\end{equation}

\noindent Finally, the cell state and the hidden state for the current time step are calculated as:
		\begin{equation}
c_t=f_t \times c_{t-1} + i_t \times \tilde{c}_t
		\label{eq_fwdcell}
		\end{equation}
		
		\begin{equation}
h_t=o_t \times tanh(c_t)
		\label{eq_fwdhid}
		\end{equation}

\noindent Note that, in the above equations, each state will be updated by the same input vector $x_{t}$ and hidden state vector $h_{t-1}$. Equation \ref{eq_fwdsi}-\ref{eq_fwdsc} can then be represented by a single GEMM as described in equation \ref{eq_fwdmerge}:
		\begin{equation}
\begin{bmatrix} si_t \\ sf_t \\ so_t \\ s\tilde{c}_t \end{bmatrix} = \begin{bmatrix} W_i \\ W_f \\ W_o \\ W_c \end{bmatrix}x_t + \begin{bmatrix} R_i \\ R_f \\ R_o \\ R_c \end{bmatrix}h_{t-1} 
		\label{eq_fwdmerge}
		\end{equation}

\noindent where $si_{t}$, $sf_{t}$, $so_{t}$, $s\tilde{c}_{t}$ form a large tensor constructed by concatenating the individual buffers. Since the input vectors at different time steps are independent of each other, the computations for all time steps can be further fused in a single GEMM call as illustrated in equation \ref{eq_fwdingemm} below. 

		\begin{equation}
\begin{bmatrix} s_0 & s_1 & \dots & s_{T-1} \end{bmatrix} = W\begin{bmatrix} x_0 & x_1 & \dots & x_{T-1} \end{bmatrix}
		\label{eq_fwdingemm}
		\end{equation}	

\noindent where
		\begin{equation}
s_t=\begin{bmatrix} si_t \\ sf_t \\ so_t \\ s\tilde{c}_t \end{bmatrix}
		\label{eq_fwdingemm_s}
		\end{equation}
\noindent and
		\begin{equation}
W=\begin{bmatrix} W_i \\ W_f \\ W_o \\ W_c \end{bmatrix}
		\label{eq_fwdingemm_w}
		\end{equation}
\noindent For an input sequence of length T, the above optimization yields the following advantages (1) input weight matrices $W$ for all four gates only need to be loaded once, for $x_t$ over all time steps, leading to (T-1) savings in memory reading of the four matrices; (2) at each time step, the number of loads for the hidden state vector at the previous time step $h_{t-1}$ in each neuron is reduced from four to one. Meanwhile, the operations in equations \ref{eq_fwdi}-\ref{eq_fwdo}  are also fused into one call of the sigmoid kernel due to the computational homogeneity and contiguous memory-layout.

\begin{figure}
	\includegraphics[scale=0.3]{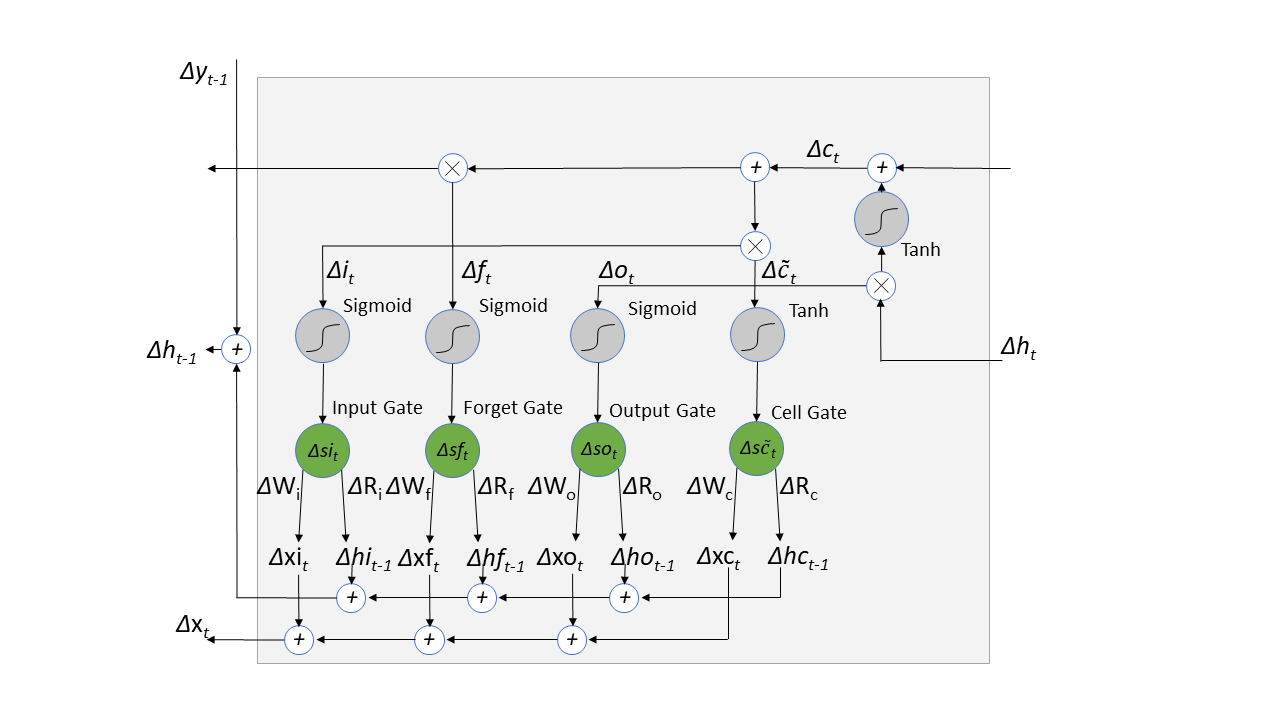}
	\caption{backward flow of LSTM}
	\label{fig_lstm_bwd}
\end{figure}

The backward path illustrated in Figure \ref{fig_lstm_bwd} adopts a similar optimization strategy. After deriving the back-propagation error $\Delta si_{t}$, $\Delta sf_{t}$, $\Delta so_{t}$, $\Delta s\tilde{c}_{t}$ at all four gates, the error propagating to the previous state can be derived as depicted in equation \ref{eq_bwdmerge}:

		\begin{equation}
\Delta h_{t-1}=\begin{bmatrix} R_i^T & R_f^T & R_o^T & R_c^T \end{bmatrix}\begin{bmatrix} \Delta si_t \\ \Delta sf_t \\ \Delta so_t \\ \Delta s\tilde{c}_t \end{bmatrix} + \Delta y_{t-1}
		\label{eq_bwdmerge}
		\end{equation}

\noindent while $\Delta y_{t-1}$ is the error propagated from the higher stack of LSTM layer and can be populated in $\Delta h_{t-1}$ buffer beforehand. After updating state errors of each gate over all time, a single GEMM call yields the back-propagation error $\Delta x_t$ for the lower LSTM layer as shown in equations \ref{eq_bwdinmerge} and \ref{eq_bwdingemm}.

		\begin{equation}
\Delta x_t=\begin{bmatrix} W_i^T & W_f^T & W_o^T & W_c^T \end{bmatrix}\begin{bmatrix} \Delta si_t \\ \Delta sf_t \\ \Delta so_t \\ \Delta s\tilde{c}_t \end{bmatrix}
		\label{eq_bwdinmerge}
		\end{equation}	

		\begin{equation}
		 \begin{aligned}
\begin{bmatrix} \Delta x_0 & \Delta x_1 & \dots & \Delta x_{T-1} \end{bmatrix} \\
= W^T \begin{bmatrix} \Delta s_0 & \Delta s_1 & \dots & \Delta s_{T-1} \end{bmatrix}
		 \end{aligned}
		\label{eq_bwdingemm}
		\end{equation}	

\noindent In the ideal case, the backpropagation error for weights may be updated using two GEMM calls if the batch sizes are the same for all time steps.
The input weight update at each time step is given by:

		\begin{equation}
		 \begin{aligned}
\begin{bmatrix} \Delta W_i & \Delta W_f & \Delta W_o & \Delta W_c \end{bmatrix} \\
= \begin{bmatrix} \Delta si_t & \Delta sf_t & \Delta so_t  & \Delta s\tilde{c}_t \end{bmatrix} x_t^T
\end{aligned}
		\label{eq_bwdinwei}
		\end{equation}

\noindent In MIOpen, the input weight update over all time steps is given by a single GEMM call as in equation \ref{eq_bwdinwei1}:

		\begin{equation}
\Delta W = \begin{bmatrix} \Delta s_0 & \Delta s_1 & \dots & \Delta s_{T-1} \end{bmatrix} \begin{bmatrix} x_0^T \\ x_1^T \\ \dots \\ x_{T-1}^T \end{bmatrix}
		\label{eq_bwdinwei1}
		\end{equation}

\noindent The hidden state weight update at each time step is given by:

		\begin{equation}
		 \begin{aligned}
\begin{bmatrix} \Delta R_i & \Delta R_f & \Delta R_o & \Delta R_c \end{bmatrix} \\
= \begin{bmatrix} \Delta si_t & \Delta sf_t & \Delta so_t  & \Delta s\tilde{c}_t \end{bmatrix} h_{t-1}^T
\end{aligned}
		\label{eq_bwdhidwei}
		\end{equation}	

\noindent Similar to the single-GEMM call for input weight update, the hidden state weight update over all time is

		\begin{equation}
\Delta R = \begin{bmatrix} \Delta s_0 & \Delta s_1 & \dots & \Delta s_{T-1} \end{bmatrix} \begin{bmatrix} h_{-1}^T \\ h_0^T \\ \dots \\ h_{T-2}^T \end{bmatrix}
		\label{eq_bwdhidwei1}
		\end{equation}

\noindent In parallel computing, data tensors are usually packed in batches. However, when training different lengths of sentences in an LSTM model, the batch size at each time step can be different. MIOpen requires a length-descending arrangement for batched sentences (longest sentence at the top of the batch while the shortest at the bottom) to guarantee computational efficiency. In practice, consistent batch size along time axis is preferred to achieve the best performance. For instance, in backward weight update, if the batch size varies along the time axis, the GEMM results of hidden state vectors will have to be aligned at each time step and then subsequently accumulated. This will result in $T + 1$ separate GEMM calls in total, instead of just $2$. 

\subsection{Other Primitives}
In addition to the operations above, other operations are required to support the bulk of the computational workload in popular neural network architectures. Among these operations MIOpen implements the following operations for both training and inference:

\begin{enumerate}
	\item Activation Operations
	\item Pooling 
	\item Softmax
	\item CTC Loss Function
	\item Tensor Operators
	\item Local response normalization
\end{enumerate}

The procedure to invoke these operations is similar to convolution with the exception that they do not require the \emph{find step}.

\section{Fusion API}
\label{sec_fusion}
Most neural networks are data-flow graphs where data flows from one direction and is operated upon as it moves from one layer to another. While conceptually data is flowing only in one direction, the underlying kernels implementing these operations have to read data from the global memory, operate on the data and then write the result back for layers down the pipeline. This is necessary due to the limited on-chip memory of the GPUs given the large image and filter sizes in neural network architectures. 

However, not all operations require that data be read from and written back to the global memory each time. That is some operations may be fused to increase the compute efficiency of these kernels. This merger of the operations to be performed by a single kernel may be termed as \emph{kernel-fusion}.

As a simple example let's consider an addition operation followed by a rectified linear unit (ReLU) operation. In this case, the intermediate result need not be written back to the main memory, and both the operations may be performed while the individual data elements are in the on-chip memory. Another common sequence of operations is convolution followed by a bias (addition) and ReLU operation. It must be kept in mind that fusions for other operators are much more involved such as the fusion of the convolution and batch normalization operation.

The MIOpen library offers the fusion API to facilitate the efficient fusion of such operations; it allows the user to specify a sequence of operations that are desired to be fused. Once the user specifies this sequence, MIOpen decides the applicable kernel and compiles it; all this information is encapsulated in the \texttt{miopenFusionPlanDescriptor} data structure\cite{miopen_github}.

\begin{figure}[htbp]
	\centerline{\includegraphics[scale=0.3]{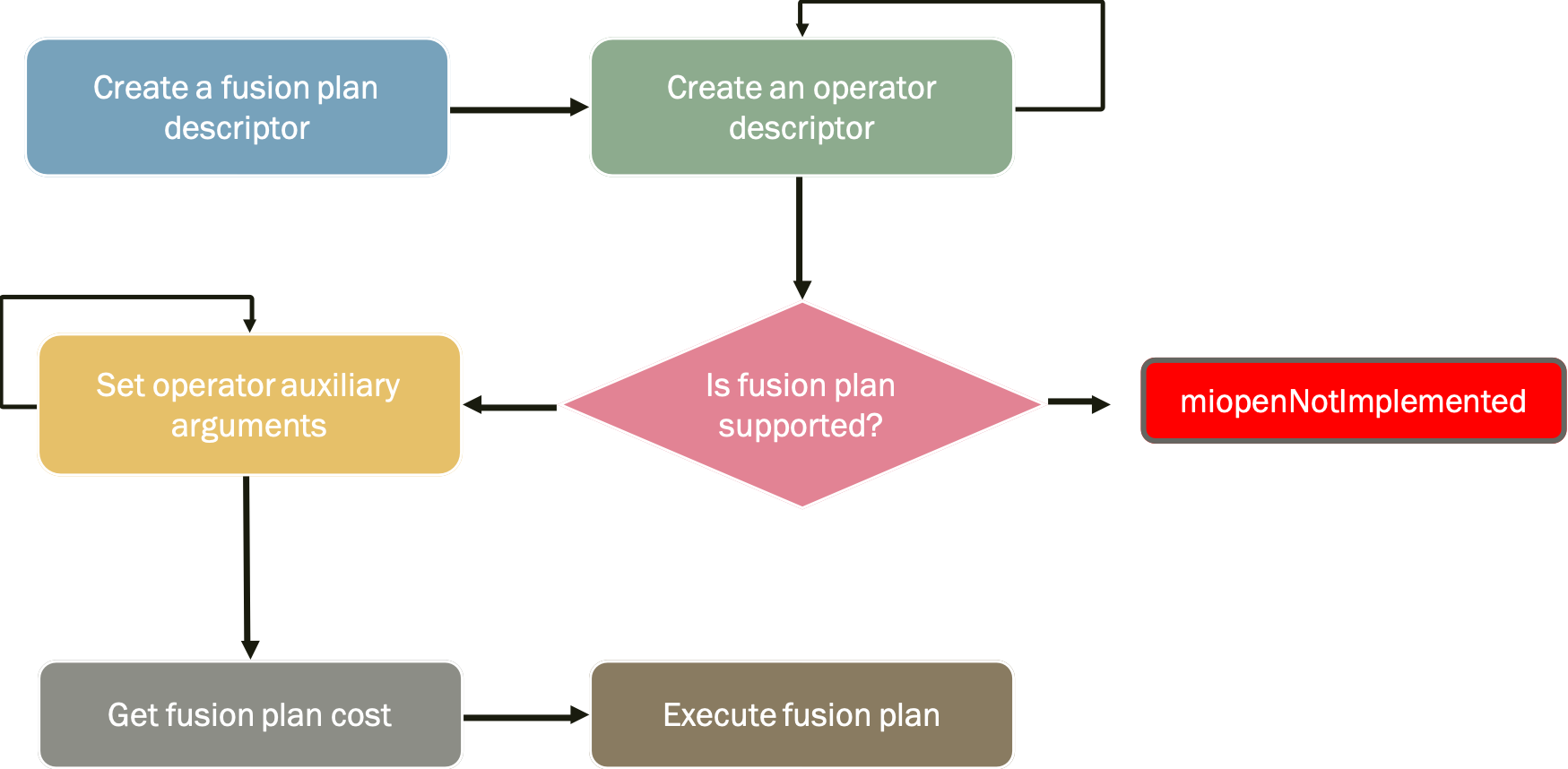}}
	\caption{Steps for creating and executing a fusion plan}
	\label{fig_fusion}
\end{figure}

If merging of the required fusion sequence is feasible, the compilation step of the fusion plan will return success; thereafter the user would supply the runtime arguments for the kernels such as parameters for different operations. Following which, the user would execute the fusion plan with data pointers for the input and output data. The advantage of separating the compilation step from the argument structure is that the fusion plan which has been compiled once, need not be compiled again for different input values. Figure \ref{fig_fusion} shows a pictorial representation of the steps required to create a fusion plan. Further details and example code can be found at \cite{miopen_doc}.

\subsection{Metadata graph}

Internally MIOpen relies on a constraint specification graph, which when traversed with the attributes of fusion operations results in the applicable kernels. Such a mechanism allows the addition of new fused kernels with an arbitrary sequence of operations without the combinatorial increase in complexity. 

\subsection{Supported Fusions}
Tables \ref{fig_fusion_supported_fp32} and \ref{fig_fusion_supported_fp16} enumerates the different combinations of fusions that are currently supported by MIOpen for single precision and half precision respectively. The first column indicates the combination of operations that may be fused, where C stands for Convolution, B for bias, N for Batch Nomralization and A for activation. 

\begin{table*}[]
\caption{Fusions supported by MIOpen (Single Precision) }
\centering
\label{fig_fusion_supported_fp32}
\begin{tabular}{lllllll}
\textbf{Combination} & \textbf{Conv Algo} & \textbf{Stride} & \textbf{Filter Dims} & \textbf{BN Mode} & \textbf{Activations} & \textbf{Other Constraints} \\
\hline
CBNA & Direct & 1 and 2 & 3x3, 5x5, 7x7, 9x9, 11x11 & All & All & stride and padding must be either 1 or 2 \\
\hline
\multirow{12}{*}{CBA} & Direct &  & 1x1 &  & All & stride/ padding not supported \\
 & \multirow{11}{*}{Winograd} & 1 & 1x1, 2x2 & N/A & Relu, Leaky Relu & c \textgreater{}= 18 \\
 &  & 1 & 3x3 &  & Relu, Leaky Relu & c \textgreater{}= 18 and c is even \\
 &  & 1 & 4x4, 5x5, 6x6 &  & Relu, Leaky Relu & 4 x c \textgreater{}= 18 \\
 &  & 1 & 7x7, 8x8, 9x9 &  & Relu, Leaky Relu & 12 x c \textgreater{}= 18 \\
 &  & 1 & 10x10, 11x11, 12x12 &  & Relu, Leaky Relu & 16 x c \textgreater{}= 18 \\
 &  & 1 & larger filter sizes &  & Relu, Leaky Relu & none \\
 &  & 2 & 1x1 &  & Relu, Leaky Relu & 2 x c \textgreater{}= 18 \\
 &  & 2 & 2x2, 3x3, 4x4, 5x5, 6x6 &  & Relu, Leaky Relu & 4 x c \textgreater{}= 18 \\
 &  & 2 & 7x7 &  & Relu, Leaky Relu & 12 x c \textgreater{}= 18 \\
 &  & 2 & 8x8, 9x9, 10x10, 11x11, 12x12 &  & Relu, Leaky Relu & 16 x c \textgreater{}= 18 \\
 &  & 2 & larger filter sizes &  & Relu, Leaky Relu & none \\
\hline
NA & - &  & - & All & All & Padding not supported
\end{tabular}%
\end{table*}

\begin{table*}[]
\caption{Fusions supported by MIOpen (Half Precision) }
\centering
\label{fig_fusion_supported_fp16}
\begin{tabular}{lllllll}
\textbf{Combination} & \textbf{Conv Algo} & \textbf{Stride} & \textbf{Filter Dims} & \textbf{BN Mode} & \textbf{Activations} & \textbf{Other Constraints} \\
\hline
CBNA & Direct & 1 and 2 & 3x3, 5x5, 7x7, 9x9, 11x11 & All & All & stride and padding must be either 1 or 2 \\
CBA & Direct &  & 1x1 &  & All & stride/ padding not supported
\end{tabular}%
\end{table*}

\section{Results}
\label{sec_stat}
This section highlights the performance improvements that MIOpen is able to offers particularly in convolution as well as some supported fusions. To date, the primary beneficiary of Machine Learning progress has been machine vision as well as Natural Language processing. In machine vision, the convolution operation is the primary workhorse due to the low number of parameters required to learn as compared to regular neural networks as well as the favorable mathematical properties. However, the parameters associated with the convolution operations in different deep convolution neural networks have changed considerably. The early CNNs employed larger filter sizes to reduce the height and width of the feature maps and simultaneously increase the number of feature maps. For instance, LeNet \cite{lecun1995convolutional} employed filters of size $5 \times 5$ while, Alexnet \cite{krizhevsky2012imagenet} contained filters of size $5\times 5$ as well as $11 \times 11$. However, recently \cite{He2016resnet}, \cite{szegedy2015going} networks have almost exclusively relied on smaller filter sizes namely only $1 \times 1$ and $3 \times 3$ coupled with striding to reduce the size of the feature map. 

\begin{figure*}
\centering
\begin{tabular}{cc}
\subfloat[1x1 filter size (Forward) \label{fig_conv_compare_1x1F}]{\includegraphics[scale=0.45]{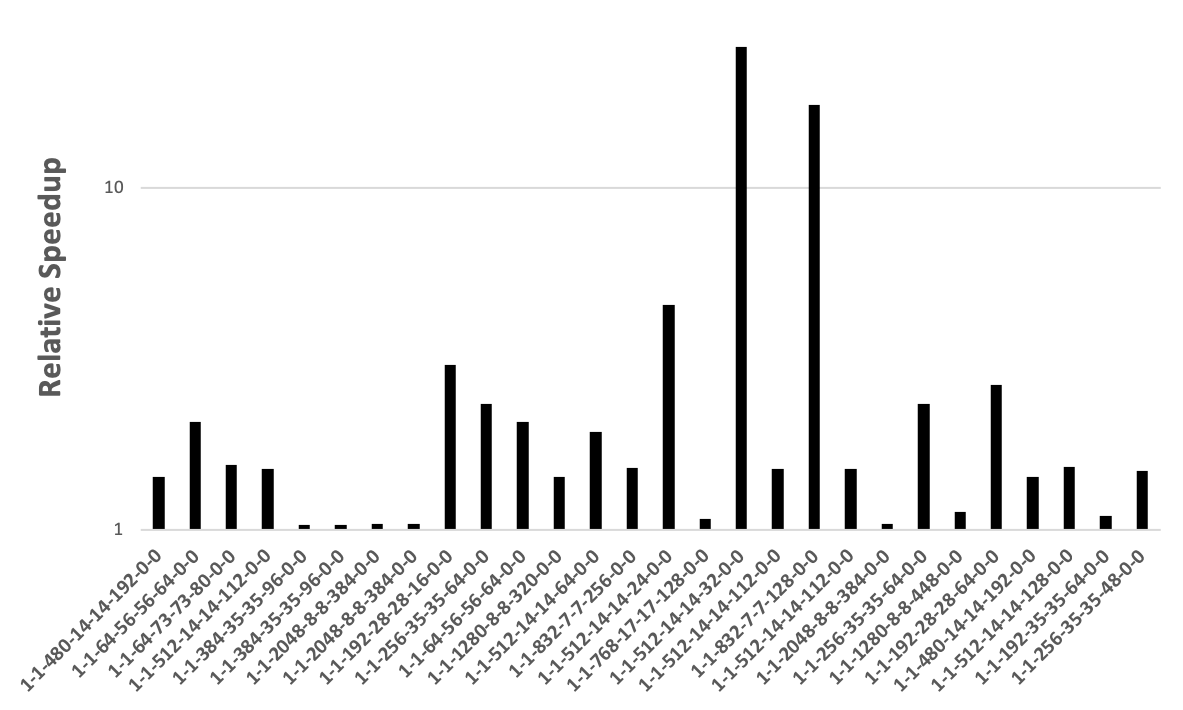}} & 
\subfloat[non 1x1 filter sizes (Forward) \label{fig_conv_compare_n1x1F}]{\includegraphics[scale=0.45]{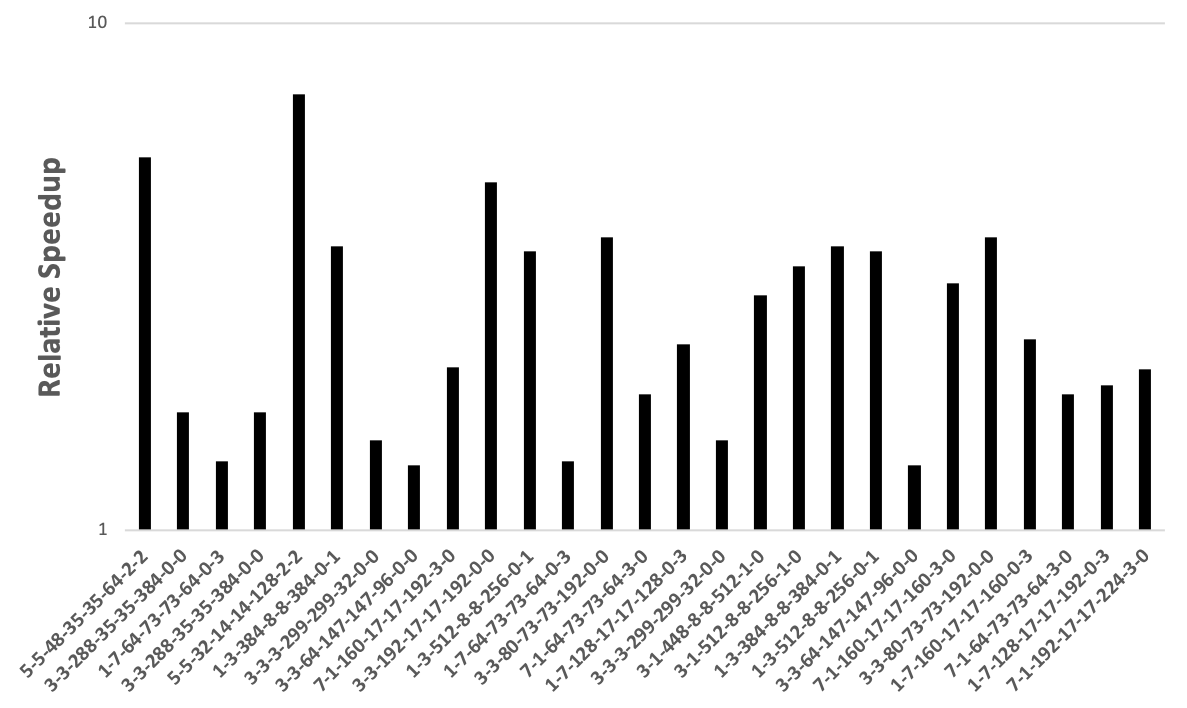}} \\
\subfloat[1x1 filter size (Backward Data)\label{fig_conv_compare_1x1B}]{\includegraphics[scale=0.45]{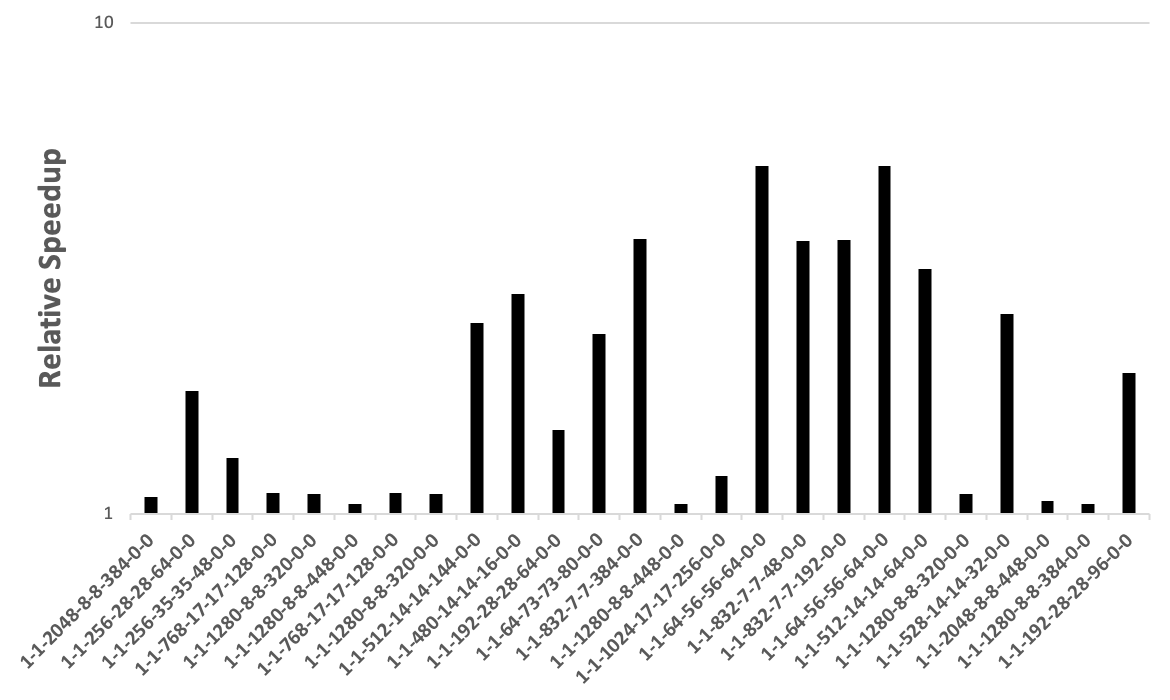}} 
&
\subfloat[non 1x1 filter sizes (Backward Data)\label{fig_conv_compare_n1x1B}]{\includegraphics[scale=0.45]{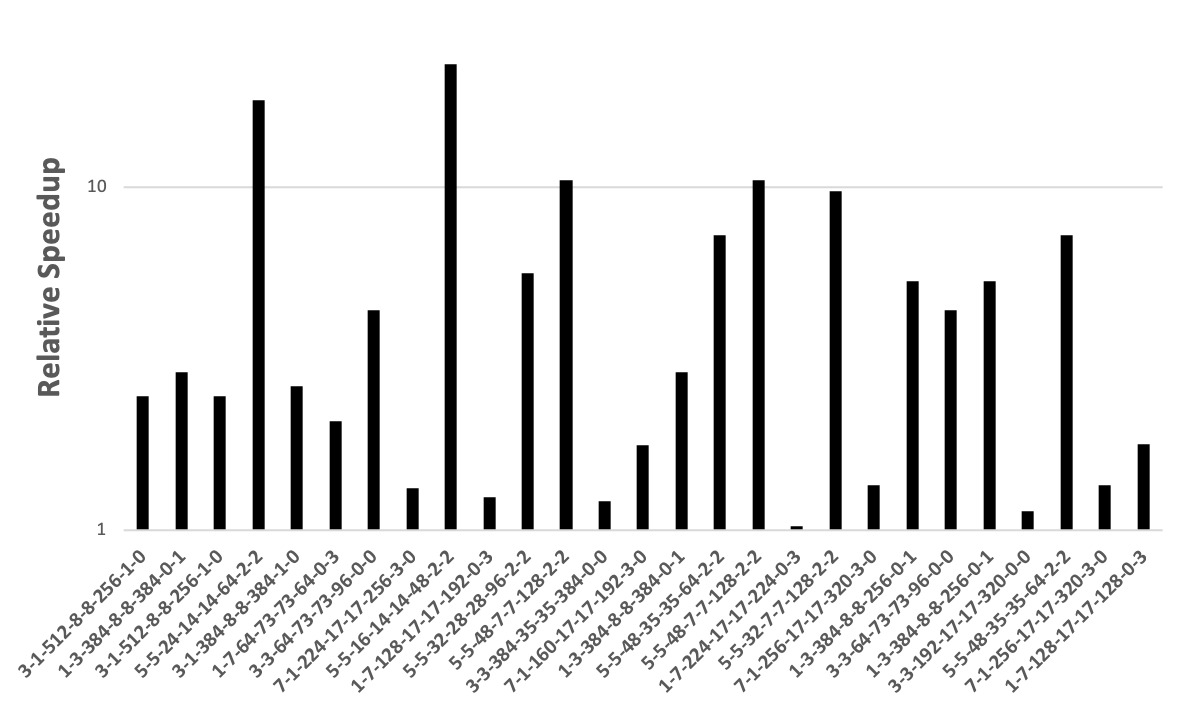}} \\
\subfloat[1x1 filter size (Backward Weights)\label{fig_conv_compare_1x1W}]{\includegraphics[scale=0.45]{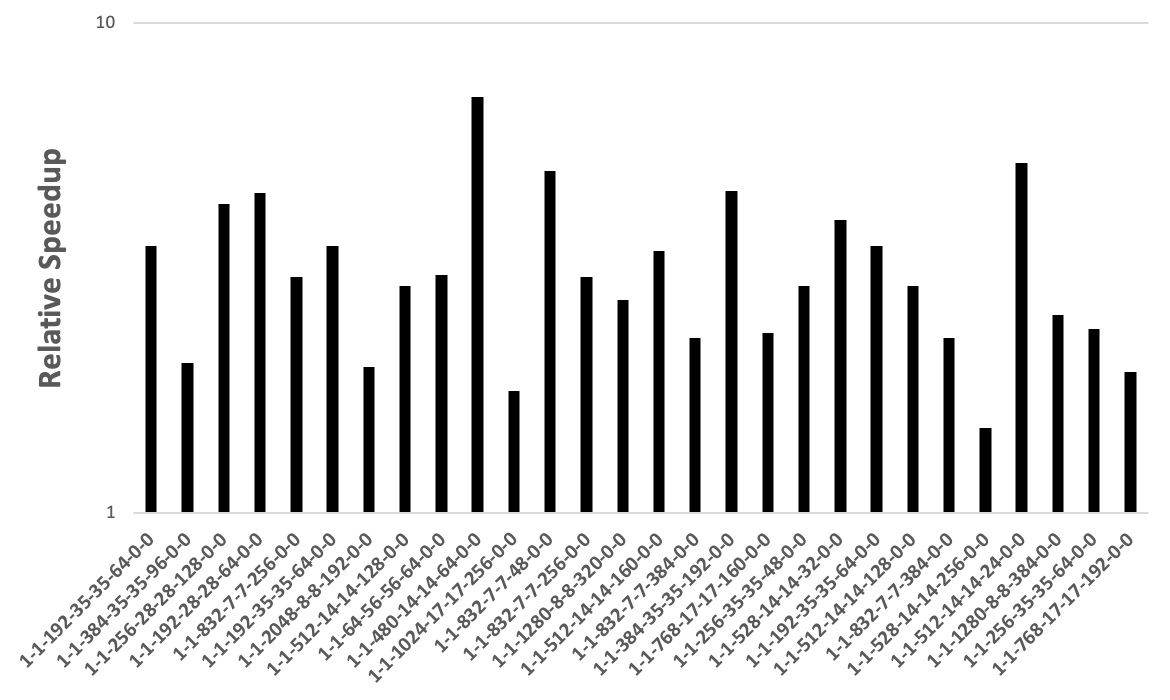}} 
&
\subfloat[non 1x1 filter sizes (Backward Weights)\label{fig_conv_compare_n1x1W}]{\includegraphics[scale=0.45]{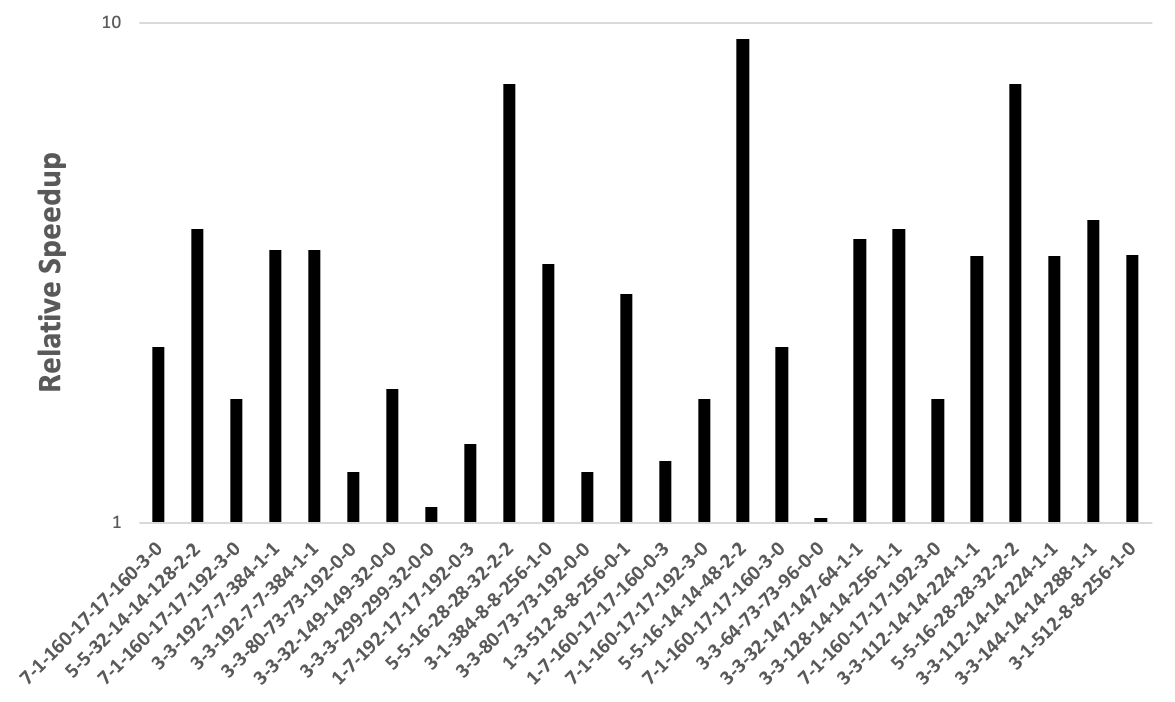}} \\
\end{tabular}
\caption{Relative performance improvement for different convolution configurations as compared to im2col+GEMM }
\label{fig:conv_compare}
\end{figure*}

Figure \ref{fig:conv_compare} shows the relative speedup of different convolution configurations as compared to MIOpen's im2col+GEMM implementation. The configurations shown therein have been selected randomly from different popular networks such as GoogLeNet, Inception v3, and Inception v4 \cite{szegedy2015going} for image classification. The y-axis in Figure \ref{fig:conv_compare} shows $\log$ of the speedup obtained by MIOpen, while the x-axis shows the labels for different configurations. Each label shows, respectively, the filter height, filter width, input channels, image height, image width, output channels, padding (height) and padding (width) separated by a hyphen (-). 

Figures \ref{fig_conv_compare_1x1F}, \ref{fig_conv_compare_1x1B} and \ref{fig_conv_compare_1x1W} depict the performance gains for kernels with filter height and width equal to 1 ($1 \times 1 $ convolutions) in the forward, backwards-data and backwards-weights directions respectively. While mathematically $1 \times 1$ convolutions may be described as a pure GEMM operation, still MIOpen may provide substantial performance benefit in certain cases. Similarly, Figures \ref{fig_conv_compare_n1x1F}, \ref{fig_conv_compare_n1x1B} and \ref{fig_conv_compare_n1x1W} show the performance benefit attained for non-$1\times1$ kernels in the forward, backward-data and backward-weights directions respectively.

As mentioned in Section \ref{sec_prims} MIOpen employs the Winograd algorithm for applicable convolutions while the $1 \times 1$ convolutions are primarily serviced by kernels written in GCN ISA. Due to the efficiency of the Winograd algorithm, MIOpen can speed up many $3 \times 3$ convolutions, however, on larger filter sizes it is not as effective due to granularity loss. Wherein MIOpen's other convolutional kernels step in to provide speedup, however, in some cases, this speedup is not substantial. The MIOpen team is continuously working on new algorithms to improve performance in these areas.

\begin{figure*}
    \centering
    \begin{tabular}{cc}
        \subfloat[Speedup with Fusing Convolution + Bias + Activation \label{fig_cba}]{\includegraphics[scale=0.45]{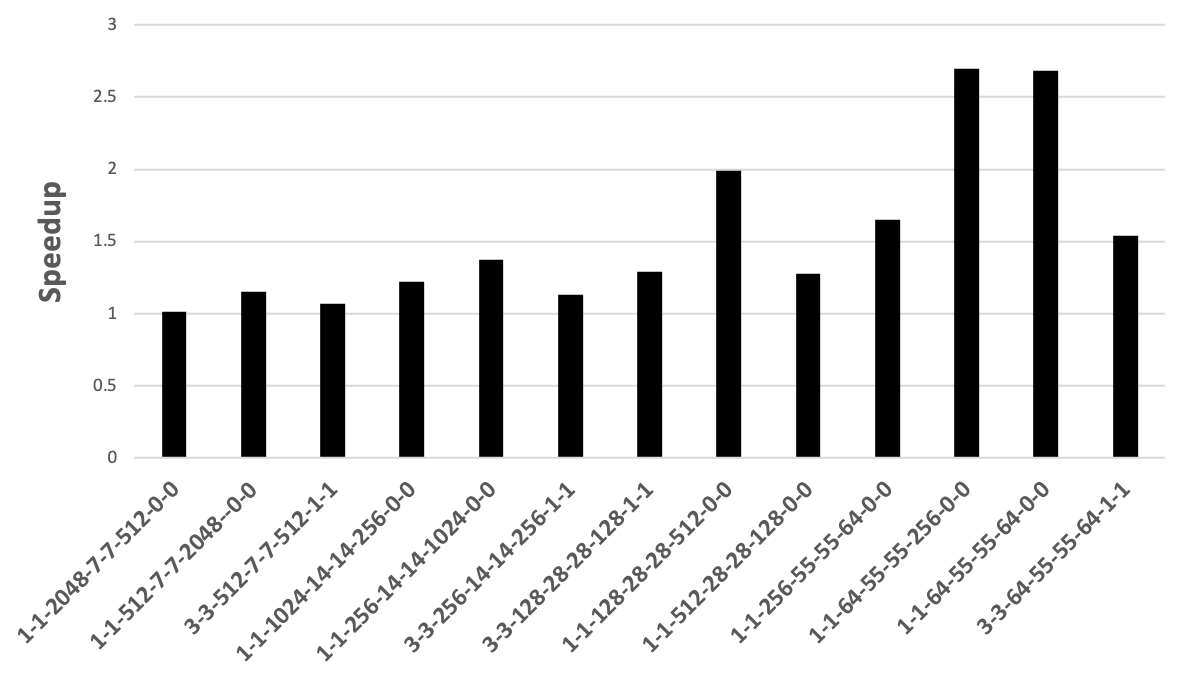}} &
        \subfloat[Speedup with Fusing Batchnorm + Activation \label{fig_na}]{\includegraphics[scale=0.45]{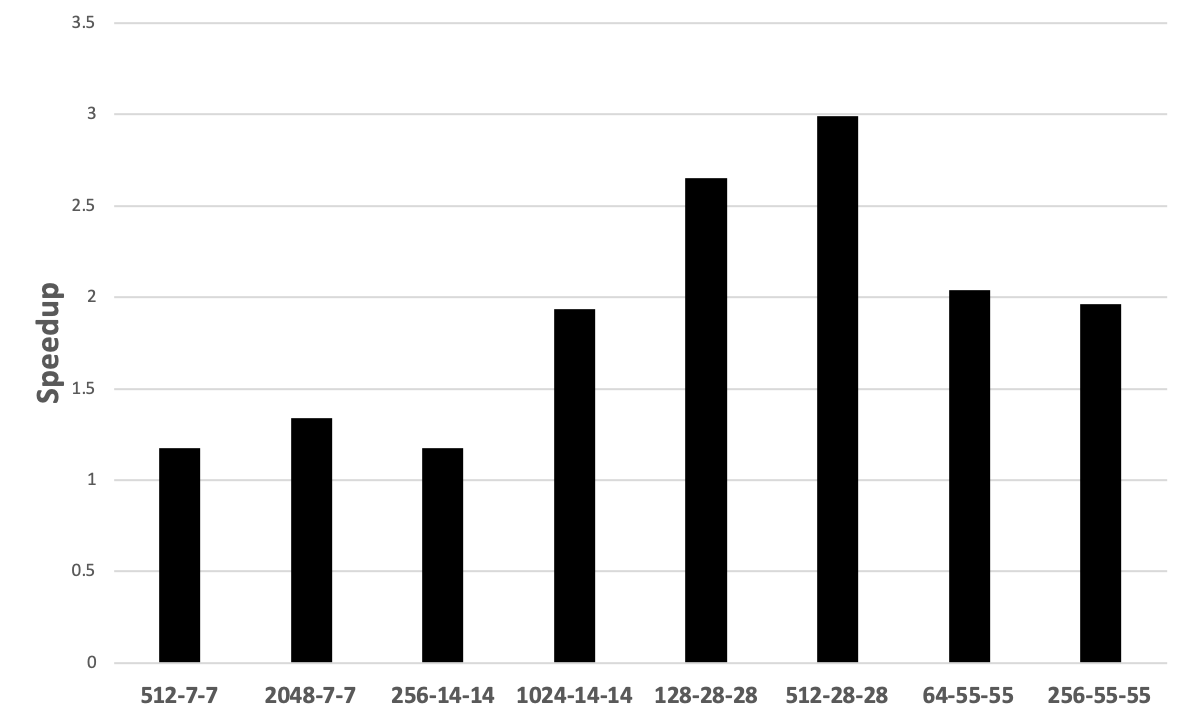}} \\
    \end{tabular}
    \caption{Relative performance improvement for different fused configurations compared to their non-fused counterparts}
    \label{fig_fusion_res}
\end{figure*}

Section \ref{sec_fusion} describes the MIOpen Fusion API, which allows the user to fuse many arbitrary combinations of operations to reduce memory traffic and provide performance gains. Figures \ref{fig_cba} and \ref{fig_na} depict the speedup achieved using the Fusion API. 

Figure \ref{fig_cba} indicates the speedup achieved by the fused operations versus the same operations performed individually. The amount of speedup achieved varies with different configurations, with some being accelerated to as high as $2.5$ times the separate run-times. It may be noted that higher speedup is achieved for kernels with fewer output features (channels) since a larger bias vector results in the memory system being the bottleneck. 

The MIOpen Fusion API is also capable of fusing the Batch-Normalization and Activation operation in the forward direction. The speedup achieved using this fusion for different configurations is depicted in Figure \ref{fig_na}, where the horizontal axis indicates the number of input channels, the height and width of the image. The results indicate that this fusion is more effective for larger image sizes with more number of channels, while smaller images are not able to benefit from the fused operations. However, the possible speedup using this fusion makes it a viable optimization venue to be explored. The MIOpen team is working on expanding the scope of the effectiveness of the Fusion API by enabling more fusions and improving the efficacy of the existing fusions.

\section{Conclusions and Future Work}
\label{sec_conc}
This paper identified some of the challenges faced by a high performance computing library and some of the mechanisms implemented in MIOpen to address these challenges were presented. The open source nature of MIOpen makes it easy for researchers and academics to experiment and implement novel solutions to these problems, the authors look forward to constructive feedback from the community. 

\section*{Acknowledgements}
The MIOpen team would like to gratefully acknowledge the valuable contributions of Alex	Lyashevsky, James Newling and the GitHub user \texttt{ghostplant} as well as the  support of the open source community.

© 2019 Advanced Micro Devices, Inc.  All rights reserved.
AMD, the AMD Arrow logo and combinations thereof are trademarks
of Advanced Micro Devices, Inc.  Other product names used in this publication are for identification purposes only and may be trademarks of their respective companies.
Python is a trademark of the Python Software Foundation, OpenCL is a trademark of Apple Inc. used by permission by Khronos Group, Inc.

\Urlmuskip=0mu plus 1mu\relax
\bibliographystyle{IEEEtran}
\bibliography{ms}

\end{document}